# Mixed pooling of seasonality for time series forecasting: An application to pallet transport data

**HYUNJI MOON**

Industrial Engineering, Seoul National University, mhj1667@gmail.com, 08826, 1 Gwanak-ro, Gwanak-gu, Seoul, Republic of Korea

**BOMI SONG***

School of Air Transport, Transportation and Logistics, Korea Aerospace University, 76 Hanggongdaehak-ro, Deogyang-gu, Goyang-City, Republic of Korea

*bmsong@kau.ac.kr

**HYEONSEOP LEE**

Computer Science, Korea Advanced Institute of Science and Technology, hyeonseop@kaist.ac.kr, 34141, 291 Daehak-ro, Yuseong-gu, Daejeon, Republic of Korea

**Abstract**

Multiple seasonal patterns play a key role in time series forecasting, especially for business time series where seasonal effects are often dramatic. Previous approaches including Fourier decomposition, exponential smoothing, and seasonal autoregressive integrated moving average (SARIMA) models do not reflect the distinct characteristics of each period in seasonal patterns. We propose a mixed hierarchical seasonality (MHS) model. Intermediate parameters for each seasonal period are first estimated, and a mixture of intermediate parameters is taken. This results in a model that automatically learns the relative importance of each seasonality and addresses the interactions between them. The model is implemented with Stan, a probabilistic language, and was compared with three existing models on a real-world dataset of pallet transport from a logistic network. Our new model achieved considerable improvements in terms of out of sample prediction error (MAPE) and predictive density (ELPD) compared to complete pooling, Fourier decomposition, and SARIMA model.

Keywords: Multiple seasonality; Hierarchical model; Time series forecasting; Pallet transport data

# 1. Introduction

Seasonality is the main component of time series, and the consideration of seasonality has become more important with the increasing frequency of time series produced in industry. For example, business time series data are collected and recorded on a much shorter interval. The number of web page users, the amount of electricity used, and the amount withdrawn from cash dispensers are representative examples of business time series data that had been previously collected on a quarterly or monthly basis, but is now aggregated weekly, daily, or even hourly (Hyndman, 2018). The emergence of high-frequency data, with time series containing closely spaced time intervals, necessitates a model capable of accurately modeling seasonal patterns of time series.

There is no consensus on modeling the seasonal effects: previous literature that applied computer intelligence methodologies to seasonal time series include Martínez et al. (2018) which developed a specialized kNN learning frame by dividing the training set based on seasonality and Pai, Lin, & Chang (2009) which developed SSVR, a support vector regression-based model which determines its season and trend with genetic algorithms and tabu search.

Modeling seasonality accurately, by concentrating on its two characteristics, periodic scale jumps and interactions between seasonal patterns is the main motivation for this paper. Business time series often contain periodic scale jumps, such as drops in trade volume over the weekend or extreme spikes due to stock adjustments immediately before the end-of-month evaluation. Model structures which enforce a shared parameter among subcategories with different characteristics are generally inflexible. Shared parameter method, complete-pooling, gives identical estimates without considering the differences (Gelman, 2006a). Likewise, previous seasonality models that fit entire seasonal periods with a single set of trend and seasonality parameters are not flexible enough to capture the extreme differences of scale between time components. For example, large scale differences, namely scale jumps, are observed between days of the week (Monday through Sunday) or days of the month s(the first day of the month to the last day of the month).

Moreover, business time series generally exhibit multiple seasonal patterns. As each seasonal pattern has distinct periods and effects, it is not trivial to design a model that can capture multiple seasonal patterns at once. The overall seasonal effect is not simply the sum of its parts. The necessity to reflect the interaction between seasonalities, weekly and monthly seasonality for example, hardens the problem.

As the frequency of data collection has increased, it has become essential to consider these two factors, because scale jumps and interactions occur compositely among multiple seasonalities. Previous approaches to multiple seasonal patterns include Fourier decomposition, exponential smoothing, and seasonal autoregressive integrated moving average (SARIMA) models. These approaches often fail to reflect the complex seasonalities (Hyndman and Athanasopoulos, 2018). Distinct characteristics of each period, such as the unique behavior of specific days of the week in business data are fully modeled.

Apart from the two specific seasonal characteristics, the increased variety of recent data also motivates our research. Previously, many time series models that "adaptively" or "automatically" learn data have been suggested, with illustrative examples including an auto-ARIMA model that searches through a range of different parameters and selects the best model and the Prophet forecasting model, which utilizes a Bayesian-based curve fitting method to learn trend and seasonality parameters. Also, many applied automatic frameworks for specific types of time series have been introduced (Gooijer and Hyndman, 2006). Most of these methods, although they greatly alleviate the burden of forecasting, have failed to adaptively detect multiple seasonal effects. Therefore, there is a need for a model that could automatically capture various seasonal patterns from the data.

We suggest a new time series forecasting framework, namely the mixed hierarchical seasonality (MHS) model. This approach makes use of the hierarchical structure of years, quarters, months, weeks, days, and hours inherent to time. Both the accuracy and interpretability of predictions are improved through this model, as discussed in greater depth below. A hierarchical model is the main theoretical basis of our model. According to Gelman (2006a), there is almost always an improvement, but to different degrees that depends on the heterogeneity of the observed data when hierarchical model is used. Through the hierarchical structure, the aforementioned two problems of seasonality can be improved. We solve the problem of scale jumps by separating parameters for each seasonal category, such as day of the week or day of the month. Each parameter is fit using the hierarchical model to achieve partial pooling. Partial pooling refers to the effect of hierarchical model and is generally used (Gelman et al., 2013). As its name suggests, pooled effects between subclusters are partial as they are implemented through shared hyperparameters, not parameters. Note that hyperparameters are parameters corresponding to parameters' priors. As for multiple seasonal patterns, we construct a mixed pooling model that estimates the final parameter distributions by taking a mixture of intermediate parameter distributions for each seasonal pattern.

There have been great needs for accurate forecasting as part of an expert system. The proposed model is the engine of the system and provides precise predictions on which subsequent decisions could be made. In particular, our model shines in the business management domain where demand time series with high seasonality patterns are observed. This is because demand time series are mainly generated by the direct influence of customers' cyclical activities, and the multiple seasonalities are relatively clear (Taylor & Letham, 2018). Demand forecasting is often performed as the basis for optimizing subsequent decisions. Therefore, if a predictive model yields analytical results in addition to the predicted results, it is possible to make more effective decisions in the next step. Seasonal effects in time series are important for decision-makers (Stevenson, 2015). For example, in the field of supply chain management, one purpose of demand forecasting is to prepare an appropriate supply. As demand forecasts become more accurate, even though many demand forecasting tasks are now automated, it is not easy to automate the entire process of supply chain management decision-making. Therefore, decision-makers often make final decisions based on analyses and forecasts provided to them. In this context, MHS which enhances the interpretability of

forecasts—especially with regard to seasonality—contributes to efficient management; as it can help decision-makers, who are usually suppliers, to better understand the system, control sales to buyers, and properly manage inventories (Stevenson, 2015).

The remainder of this paper consists of five sections. Section 2 introduces the key concepts upon which the MHS model is based, and section 3 explains the partial pooling model and the mixed partial pooling model. In section 4, real-world pallet transport data is introduced and experimental models are described. Section 5 contains an analysis of the experimental models, and lastly, conclusions are presented in section 6.

## 2. Literature Review

*2.1.* Past approaches to multiseasonal time series

Multiple seasonalities are discovered in diverse industries. Moreover, multiple seasonal patterns in time series vary widely across industries, especially for demand time series. For example, in the clothing industry (e.g., swimwear), the quarterly periodicity is more pronounced than the daily periodicity, and the weekly periodicity in the food industry (e.g., bakeries) is more influential than the monthly periodicity. Therefore, to reflect multiple seasonalities, it is first necessary to decompose the combined seasonal effect into each seasonal effect corresponding to every possible season. Then interactions between each season should be considered.

Among traditional statistical models, only a few models including X11, seasonal extraction in ARIMA time series (SEATS), seasonal and trend decomposition using Loess (STL), and exponential smoothing can model multiple seasonalities; X11 and SEATS methods for time series decomposition have the disadvantage of only reflecting monthly and quarterly periodicity (Dagum and Bianconcini, 2016). In contrast, STL can reflect various and time-varying periodic components (Cleveland et al., 1990), but still has the disadvantage of being applicable only when a time series is in a decomposable form (Hyndman and Athanasopoulos, 2018). Mulitseasonal approaches based on exponential smoothing include Gould et al. (2008) and Taylor (2010). However, though exponential smoothing could be effective for time series with short cycles of seasonalities (quarterly with 4 cycles or weekly with 7 cycles), it might be difficult to expect accurate forecasts for longer cycles of seasonality. For example, for data collected in 30-minute increments with daily, weekly and monthly seasonality, daily seasonal effect which requires 48 cycle period is highly complicated to be modeled due to its structural limitation (Hyndman and Athanasopoulos, 2018).

Based on these limitations, Taylor and Letham (2018) proposed a prophet model which models seasonality with Fourier regression. Fourier regression is a popular method for modeling single or multiple seasonality which models time series with Fourier terms $X_t$, regression parameter $\beta$, a regular period of the seasonality P, and the order of the Fourier series $n$, as below:

$$y(t) \sim N(k_i\, t + Xt \cdot \beta + m_i, \sigma)$$
$$X_t = [\frac{\cos(2\pi t)}{P}, \frac{\sin(2\pi t)}{P}, \ldots, \frac{\cos(2\pi n t)}{P}, \frac{\sin(2\pi n t)}{P}]$$

They compared the performance with previous models including ARIMA and exponential smoothing. They suggested that ARIMA is not effective for detecting multiple time series and that Fourier regression can detect multiple seasonality better than the exponential smoothing. In this regard, we have used Fourier regression model as the baseline and compared the forecasting accuracy with our proposed model, MHS. Note that as the purpose of our model is to suggest interpretable models that could reflect the prior knowledge on the degree of seasonality, neural network models which are known for their low ease of use and require tuning efforts (Weytjens et al., 2019) are not compared in this study.

*2.2.* Cluster-then-predict models

Cluster-then-predict forecasting is a method of creating independent prediction models for each cluster after clustering the population. Bertsimas et al. (2008) explained that since the health patterns of patients leading to heart failure vary widely, it is more accurate to group the patients' data first and then create a model based on each cluster after collecting similar patients. Barragan et al. (2016) and Kuo and Li (2016) used a cluster-then-predict framework based on the features extracted from wavelet transform. However, seasonal features, which are salient especially in business time series data, have not been included in the resulting features. Moreover, a separate model was constructed for each cluster, which could be inflexible under certain circumstances including mixture data generating processes (Gelman, 2014).

To the best to our knowledge, few approaches included a season-based feature for cluster-then-predict framework. Venkatesh et al. (2013) clustered the automatic withdrawal period of regions using a time series pattern and then made independent predictions for each community. The clustering criterion used by Venkatesh was daily cash withdrawal, grouped by the sequence-alignment method (SAM) with a sequence of length 7, which discretized the day-of-the-week effect. This method yielded more accurate predictions than the previous method of creating predictive models, without clustering preprocessing. Based on the clustering results, operators can make cash replenishment plans more efficiently. Moon and Song (2019) attempted clustering of time series data based on the seasonality features, days of the week for example. The optimal number of clusters was selected based on the goodness of fit, then based on the result, k-means clustering was performed. When applied to pallet movement amount data, a substantial increase in forecast accuracy was shown. These two approaches, however, still possess the inflexibility problem resulting from the separate models of each cluster.

The cluster-then-predict model provides a background for the MHS, as discrete models should be constructed for data with different traits. However, clustering could be too inflexible in some cases, because it does not allow for 'partial' divisions. This inflexibility could be improved through the concept of partial pooling in a hierarchical model.

*2.3.* Hierarchical model

Gelman et al. (2005) explained that hierarchical models are highly predictive because of partial pooling. When updating the model parameters, such as prior parameters, the relationship between the part of the data being used and the whole population should always be considered. If the focus is limited to the part, over-fitting occurs, whereas when the focus is only on the whole, under-fitting occurs. In a Bayesian hierarchy, the balance of fit can be learned by using hyperpriors. By properly setting the hyperprior structure, we can find a reasonable balance between over-fitting and under-fitting, as hyperpriors are known to serve as a regularizing factor. Many examples of applying hierarchical structure in cross-sectional data exist, including the eight-schools and rat tumor experiments. The structure of cross-sectional data where the whole population is divided into multiple and nested subcategories provides an excellent environment for a hierarchical model. A specific example is furnished by the structure of the eight-schools experiment, where one class belongs to one school and one school belongs to one city.

In the time series domain, the term "hierarchical time series" is not necessarily related to partial pooling. Hyndman and Athanasopoulos (2018) describe a "hierarchical time series" as time series data that can be decomposed and combined according to a category. The decomposition is often caused primarily by geographical differences, and each category may be part of a larger category so that a bundle of time series data has a hierarchical group structure. Examples include decomposing the number of Australian tourists by city and selling bicycles according to product types, such as mountain bicycles, general bicycles, and infant bicycles. Moreover, the concept of grouped time series was developed to explain that the hierarchical structure of the time series is not unique. Taieb et al. (2017) presented various methods of combining these structures harmoniously to reach a final forecast. Hyndman et al. (2011) introduced optimal decomposition and aggregation in a time series hierarchy.

*2.4.* Model evaluation measures

Information criteria can be used to measure the fit of a model. To estimate pointwise out-of-sample prediction accuracy in a Bayesian model, the widely applicable information criterion (WAIC) and the leave-one-out cross-validation (LOOCV) are preferred to the Akaike information criterion (AIC) and deviance information criterion (DIC). Due to computational problems, approximate LOOCV methods exist, including Pareto smoothed importance sampling (PSIS-LOO), which is implemented in an R package called *loo* (Vehtari et al., 2017). *loo* package provides means for both improvement and comparison. Models could first be improved with the provided diagnostic: estimation of Pareto tail shape parameter. Then they are compared with approximated expected log pointwise predictive density (ELPD). Vehtari and Lampinen (2002) suggest cross validation should be used when the estimated Pareto tail shape parameter is bigger than 0.7.

Time series cross-validation can be used to measure forecast accuracy in time series (Hyndman and Athanasopoulos, 2018). Several sets of training and test data are created in a walk-forward mode, and forecast accuracy is computed by averaging the results from multiple test sets. Various measures of forecast error exist, including the mean absolute, root mean squared, and mean absolute percentage error. For comparison between datasets with large scale differences scaled error measure is recommended but in most cases mean absolute error is recommended as it gives more stable results (Hyndman & Koehler, 2006).

### 3. Partial and Mixed pooling Model

Previous hierarchical models were mainly attempted on cross-sectional data, and no studies have focused on a hierarchical structure in time series data with multiple seasonality. However, seasonal (or recurring) characteristics of days of the week and month can be regarded as parallels to subcategories in cross-sectional data, and a considerable increase in accuracy is expected when the concept of a hierarchy is applied to model multiple seasonal patterns in time series data. The same assertion made in many previous studies (Shor et al., 2007; Tabandeh and Gardoni, 2015; Mackey et al., 2017) regarding the effectiveness of cross-sectional hierarchical models can therefore be applied to the time series domain. Therefore, in this paper, we focus on modeling the seasonal subcategory of time series using a hierarchical model. Two previously suggested problems of original seasonal models (shared trend and seasonal parameters for each seasonal subcategory and failure to model interactions between the week and month) can be improved respectively by using a partial and mixed pooling framework.

To introduce the concept of MHS, three models of seasonality are compared. These models focus on how to manage seasonal parameters in terms of pooling, as follows: complete, partial, and mixed. Both partial pooling and mixed pooling models are newly suggested. The former addresses the problem of periodic scales. However partial pooling model is unable to adapt to specific seasonalities and their interactions. Thus, mixed pooling model is further developed to improve the multiple seasonlity problem. Complete pooling model, which is an existing model, is introduced as a comparison to the partial and mixed pooling model. Stan, a platform for Bayesian statistical modeling, is used to implement the model. Stan uses effective Hamiltonian Monte Carlo sampling for parameter inference (Carpenter et al. 2017). For the Stan code of the entire model, see Appendix A.

In addition, the results are compared with those obtained using a Fourier decomposition model, which is a popular method for modeling single or multiple seasonalities that is used by Prophet. The classical time series decomposition of data into trend, seasonal, and irregular components is used. Our base model assumes a time series with a linear trend. It models the value y(t) at time $t$ as follows:

$$y(t) = kt + m + \epsilon_t \tag{1}$$

where $k$ is the growth rate, $m$ is the offset parameter, and $\epsilon_t$ is the error term.

### 3.1. Complete pooling model

In the complete pooling model, all data points share a single set of parameters $k$ and $m$. It can be represented as:

$$y(t) \sim N(kt + m, \sigma) \tag{2}$$

Entire seasonal periods are fit with a single set of parameters in complete pooling. Extreme seasonal patterns, such as periodic scale jumps, are hardly captured. This observation leads to the suggestion of a partial pooling model.

### 3.2. Partial pooling model

In a partial pooling model, we group data points by periods in seasonality, such as day of the week or day of the month. Each period in seasonality has its own set of parameters $k_i$ and $m_i$, and those parameters share a single set of priors $N(\mu_k, \sigma_k)$ and $N(\mu_m, \sigma_m)$. $\mu_k$ and $\sigma_k$ are hyperparameters. This forms a hierarchical model where parameters share information through their priors. This structure can be represented as:

$$k_i \sim N(\mu_k, \sigma_k) \tag{3a}$$
$$m_i \sim N(\mu_m, \sigma_m) \tag{3b}$$
$$Y = (y_1, \ldots, y_t)' \sim N(K_i t + M_i, \sigma) \tag{3b}$$

Just as different hierarchical structures exist in cross-sectional data, time series data could also be represented with more than one structure. For daily data, pooling could be done between parameters reflecting days of the week, days of the month, or even both. For pooling based on days of the week, the total data is divided based on the day of the week (from Monday to Sunday) into 7 submodels, while 31 submodels are created for pooling based on the day of the month. Note that considering the months with 28, 29, or 30 days, only the available day of the months are used for each model. For example, only the 31st day of the month is used for the 31th submodel.

However, in reality, certain days reflect the result of mixed seasonal effects of both days of the week and days of the month; the effect of multiple seasonalities is not the simple sum of each seasonal effect in this case. Interactions exist, leading to the last model – the mixed pooling model. This process is described in Fig. 1.

*3.3. Mixed pooling model*

The partial pooling model provides a distinct set of parameters for each seasonal period. However, it becomes unclear how to propose a likelihood for an observed value $y$ when we model a time series with more than one seasonal pattern. In the mixed pooling model, we take the linear mixture of each parameter to obtain a single set of parameters. Note that other forms of mixture, such as nonlinear, could be used. However, we used the basic linear mixture to focus on the feasibility of the model and to limit the scope of the paper. The mixture weight $\theta$ represents the contribution of each seasonal pattern to the time series. It models $k_{ij}$ $m_{ij}, y_t$ at time $t$ as follows:

$$k_{i_j} \sim N(\mu_k, \sigma_k) \tag{4a}$$

$$m_{i_j} \sim N(\mu_m, \sigma_m) \tag{4b}$$

$$Y = (y_1, \ldots, y_t)' \sim N((K_i * \theta)t + (M_i * \theta), \sigma) \tag{4c}$$

Season index is represented with index i. For example, if weekly and monthly seasonalities are considered, i = 1 corresponds to week and i = 2 corresponds to month. Index for subcategories of each season is noted with index j. For the week, there are seven subcategories in one week which leads to $1_j$ from $1_1$ to $1_7$. On the other hand, as up to 31 days exist in one month, $2_j$ is from $2_1$ to $2_{31}$. Note that in equation (4c), Y, $K_i$ and $\theta$ are vectors for each category corresponding to $y_t$, $k_{ij}$, $m_{ij}$ and $\theta_j$.

Parameter $\theta$ can be regarded as a factor reflecting the contribution of each seasonal pattern. If the data displays stronger weekly seasonality than monthly seasonality, $\theta$ for weekly seasonality is bigger. It should be noted that $\theta$ is modeled as a simplex vector, the components of which sum to 1. Fig. 1. illustrates the process for mixed pooling and the difference between partial and mixed pooling could be observed. Note that our proposed model is able to learn seasonal contributions from data. Unlike SARIMA or Fourier decomposition model where custom models should be designed for each type of seasonality, our model address seasonalities in a unified way regardless of their number and type. This relieves the user of the effort to specify the dominant type of seasonality in advance. In mixed pooling, based on the given type of seasonalities, the relative strength of each seasonality is adaptively learned from the data.

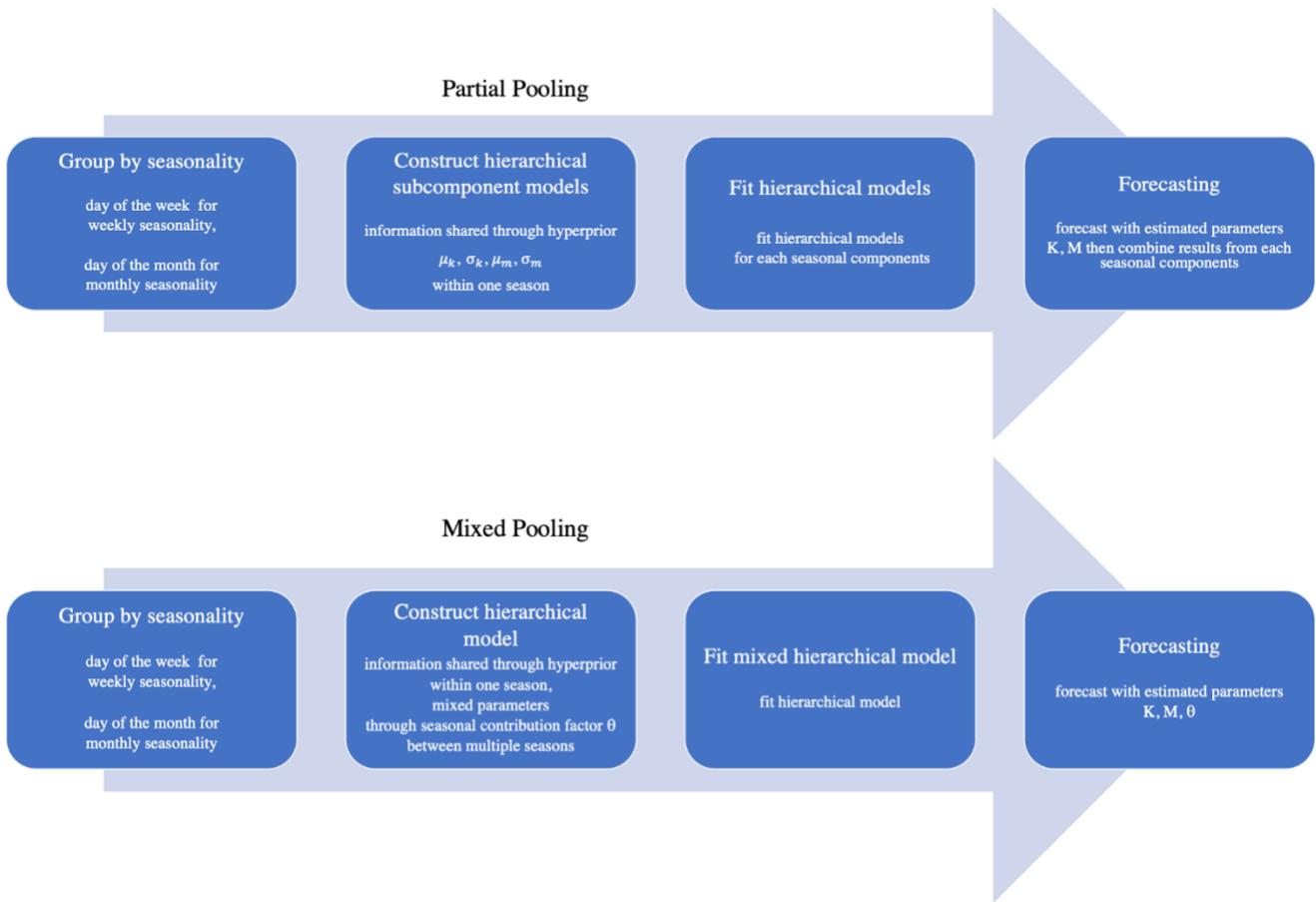

Fig. 1. Process for partial and mixed pooling

## 4. Data and Experiments

To evaluate the suggested models, we used a real-world dataset from a logistic network. Each daily time series represents the number of pallets transported in certain logistic flows: delivery, restocking, and shipment. Delivery is a flow from pallet center to factory, restocking is from factory to markets. Shipment is a returning flow from market to factory. We chose data from logistics because transport data displays strong and multiple seasonalities.

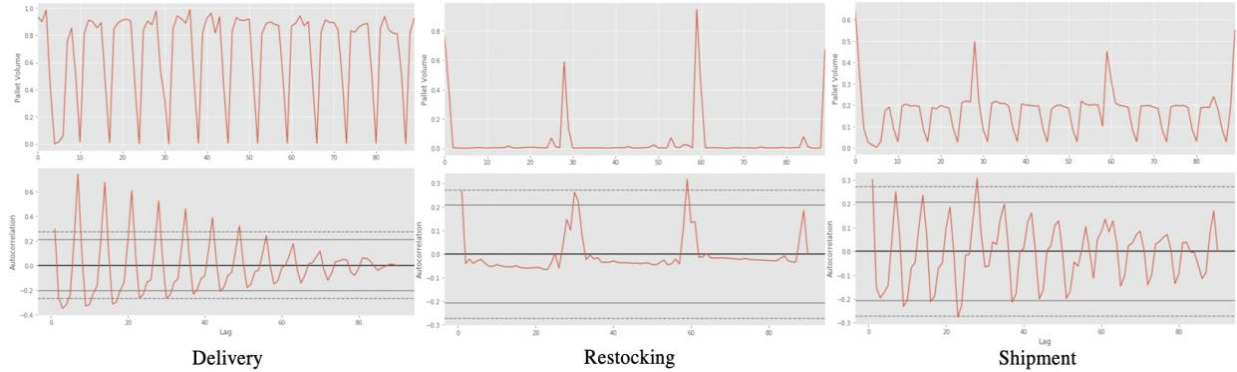

Fig. 2. Pallet volume and autocorrelations of delivery, restocking, and shipment.

Fig. 2 shows pallet volume and autocorrelation plots for each dataset. These plots reveal different seasonal patterns in each dataset. Delivery shows a strong weekly periodicity, with a sharp fall observed every Sunday. Restocking shows a clear monthly periodicity, with extreme spikes observed at the beginning and end of every month. However, its autocorrelation obscures the periodicity because the length of each month differs. While both delivery and restocking data are characterized by single seasonal periodicity, shipment shows a strong weekly periodicity as well as a subtle monthly periodicity. It implies that in the shipment data, there exists an interaction between weekly and monthly periodicity, thus the consideration of the effect of multiple seasonalities is expected to enhance the prediction performance for the shipment.

Based on the seasonal patterns observed, we constructed the suggested hierarchical models for different datasets, as shown in Fig. 3. For delivery and restocking data, partial pooling models are used as they are characterized by single seasonal periodicity. Since delivery data shows weekly seasonality and restocking data shows monthly seasonality, partial pooling models for two datasets differ in that pooling happens on days of the week for the former and days of the month for the latter. In contrast, shipment data, which displays both weekly and monthly seasonality mixed pooling model used. Pooling happens in every given seasonality, week and month, and weight which determines how each seasonality contributes to the final forecast is adaptively learned from the data. The proposed partial and mixed pooling model is compared with other models including complete pooling, Fourier decomposition, and SARIMA model.

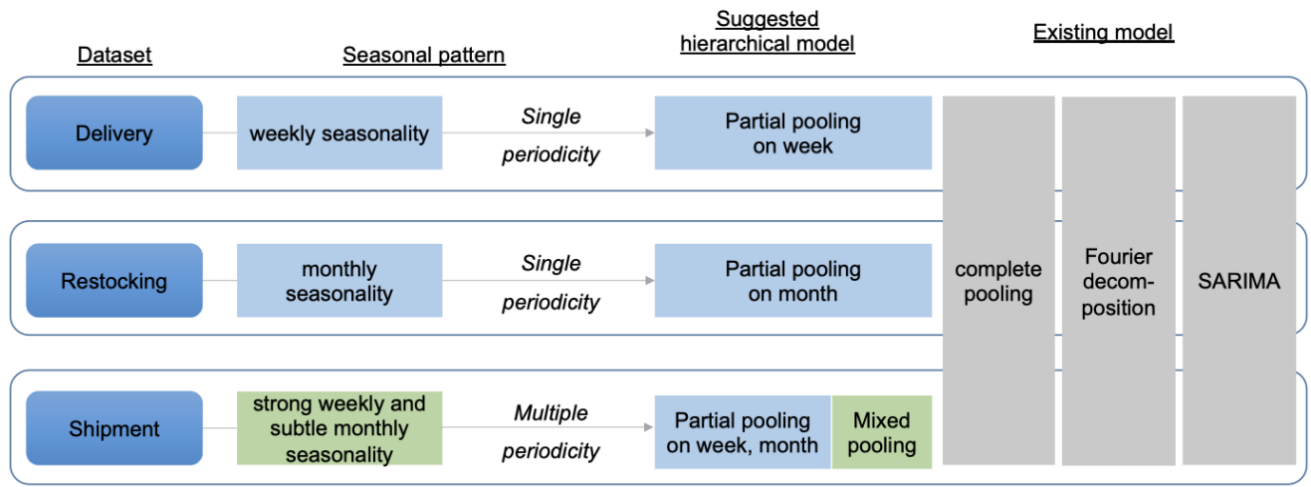

Fig. 3. Experiment design for delivery, restocking, and shipment data.

In addition, for all these three sets of data, four existing models were tested for comparison with our suggested model. First, a complete pooling model, which is an existing model was compared. Second, a Fourier decomposition model was chosen because it is a relatively recent model and has shown reportedly high performance with various seasonal data (Taylor and Letham, 2018; Usher and Dondio, 2020). Third, a SARIMA model, which is a conventional model for time series to consider seasonality was also constructed and compared. For the Fourier decomposition model, the period $P$ and Fourier order $n$ were set as 7 (days) and 3 for weekly seasonality and 30.4375 (days) and 5 for monthly seasonality. Both weekly and monthly seasonality were considered in the Fourier model for the shipment data. For SARIMA model, parameter sets (p, d, q) (P, D, Q) with the highest Bayesian information criterion (BIC) were chosen. The former three correspond to  The result was (5, 0, 3) (1, 1, 2),  (4, 0, 3) (0, 0, 1), (2, 0, 1) (0, 0, 2) with weekly, monthly, and weekly seasonality for delivery, restocking, and shipment data.

The pooling models use the Newton algorithm for robust optimization convergence. Sliding window cross-validation was used with two years of data, from 2017-01-01 to 2018-12-31. Twelve sets of test data with each of their length being 30 days were created in a sliding window manner. For example, our first test set is from 2018-01-01 to 2018-01-30 and the second test set is from 2018-01-31 to 2018-03-01. The model was fitted using all the previous data of each test set and accuracy was averaged for each test set. To measure the performance of out of sample data, mean absolute percentage error (MAPE) which is recommended for dataset with different scale (Hyndman and Athanasopoulos, 2018) and approximated expected log pointwise predictive density (ELPD) was used. ELPD-based measure is popular for Bayesian model comparison (Gelman et al., 2005).

## 5. Results and Discussion

### 5.1. Accuracy

The forecast results of the three datasets are shown in Table 1 and Fig. 3, 4, and 5.

Table 1. Comparison of out of sample error in MAPE.

| Data | Complete pooling | Fourier decomposition | SARIMA | Partial pooling | Mixed pooling |
|---|---|---|---|---|---|
| Delivery | 292.13 | 31.33 | 35.81 | 31.17 (week pool) | . |
| Restocking | 12.66 | 13.09 | 7.00 | 4.09 (month pool) | . |
| Shipment | 1.43 | 0.90 | 1.30 | . | 0.95 |

For delivery data which shows extreme scale difference, complete pooling model showed inferior forecasting performance followed by SARIMA and Fourier decomposition model (Table 1). Also, the model's inability to capture the seasonal fluctuation is presented in Fig. 3. Fourier model generally shows better performance than complete pooling or SARIMA model, the exception being restocking data. For data with single seasonality (delivery and restocking), our proposed partial pooling model displayed better ability compared to the other three models. For data with multiple seasonality, shipment, our proposed mixed pooling model showed better performance.

Table 2. Comparison of out of sample predictive density in approximated ELPD.

| Data | Complete pooling | Fourier decomposition | Partial pooling | Mixed pooling |
|---|---|---|---|---|
| Delivery | 268 | -728 | -729 | . |
| Restocking | -625 | -962 | -1409 | . |
| Shipment | -731 | -1053 | . | -1279 |

Table 2 shows the result of approximated ELPD values which are computed using Pareto-smoothed importance sampling provided from *loo* package. Lower ELPD implies a better model. Partial pooling and mixed pooling outperforms the other models for every dataset. Note that SARIMA could not be compared with this measure as it is not a Bayesian model. As the novelty in this paper lies in the new way of pooling, poolings models are compared in depth from the following plots.

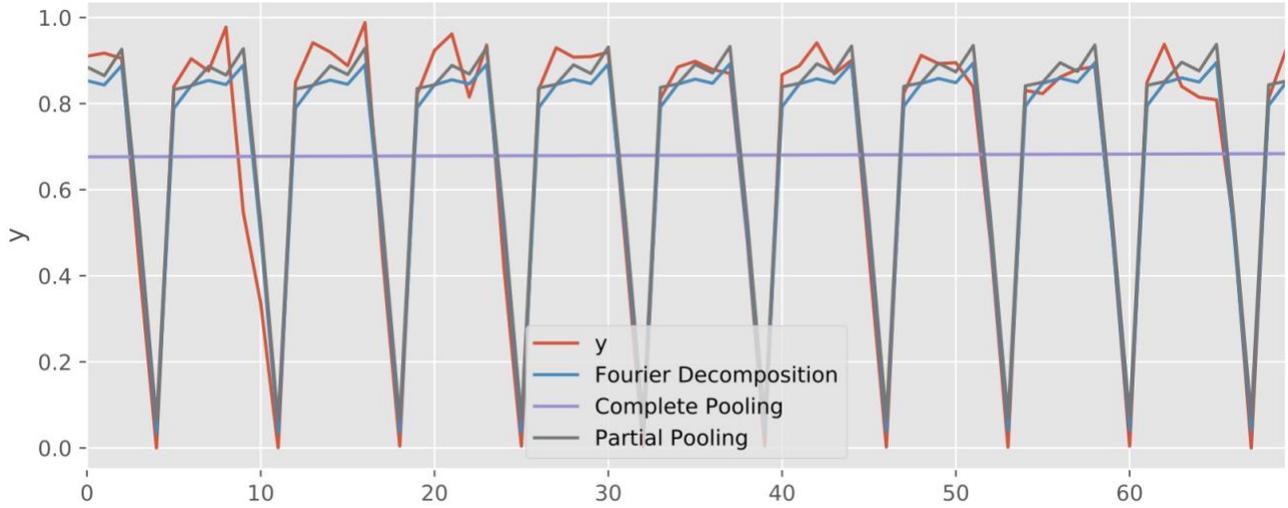

Fig. 3. Forecast for delivery data.

From Fig. 3, it can be seen that the complete pooling model does not consider weekly seasonality at all, because every data point shares a single set of parameters. The Fourier decomposition model and partial pooling model behave similarly and a negligible difference in performance is observed. In Fig. 3, 4, 5 y stands for the real pallet volume and each forecast is based on fourier decomposition, complete pooling, mixed pooling.

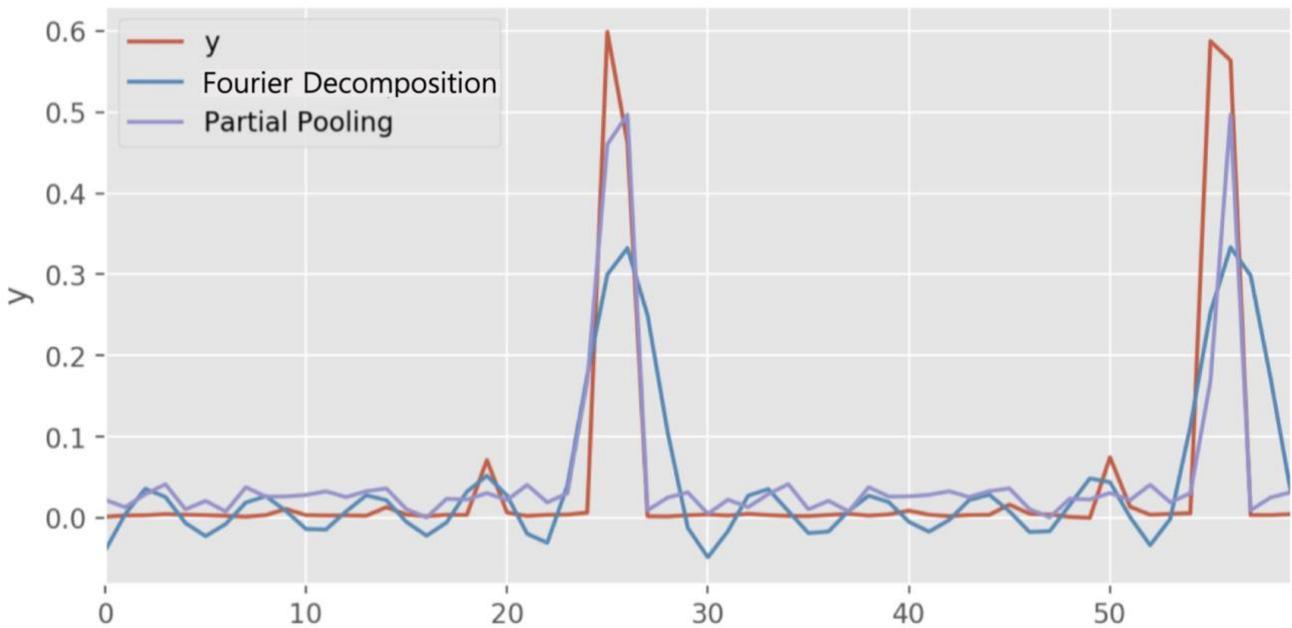

Fig. 4. Forecast for restocking data.

The mixed pooling model shows a significantly higher performance than the other models. Fig. 4 shows that the Fourier decomposition model with a given Fourier order does not fit the extreme spikes accurately. The motive for pooling model was to improve periodic scale jumps: judging from the results suggested in Table 1 and Fig. 4. the model seems to achieve its purpose.

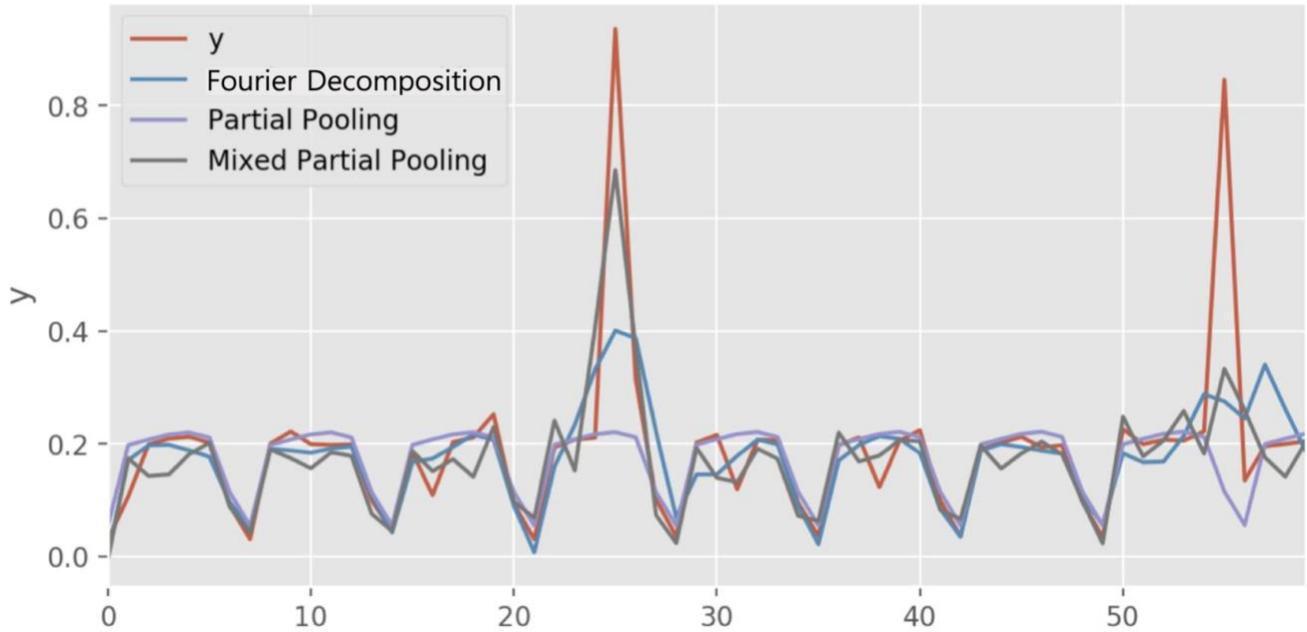

Fig. 5. Forecast for shipment data

In Fig. 5, mixed pooling model shows the highest performance. As can be seen from the figure, partial pooling model is unable to capture the multiple seasonalities. It fits on larger values influenced by monthly spikes. Fourier decomposition model performs better than partial polling but unable to fit monthly seasonality accurately similar to restocking.

*5.2. Parameters*

5.2.1. Parameter divergence in partial pooling

Fig. 6 shows trend parameters, k and m, for the delivery, restocking, and shipment datasets. In the figure for weekly seasonality parameters, number index 0 to 6 correspond to the day of the week (from Monday to Sunday), while 0 to 31 corresponds to the day of the month. In the delivery dataset, there is a sharp fall in the time series every Sunday (Fig. 2). The parameter plots from the partial pooling model in Fig. 6 show that the probability distribution of the parameters $m$ and $k$ of the sixth day of the week is differentiated from the other parameters. Similarly, in the restocking dataset, the probability distributions of the parameters $m$ and $k$ corresponding to the beginning, middle, and end of the month are differentiated. For shipment data, we have two parameter plots each corresponding to weekly and monthly parameters. Divergence of parameters is also seen from the plot.

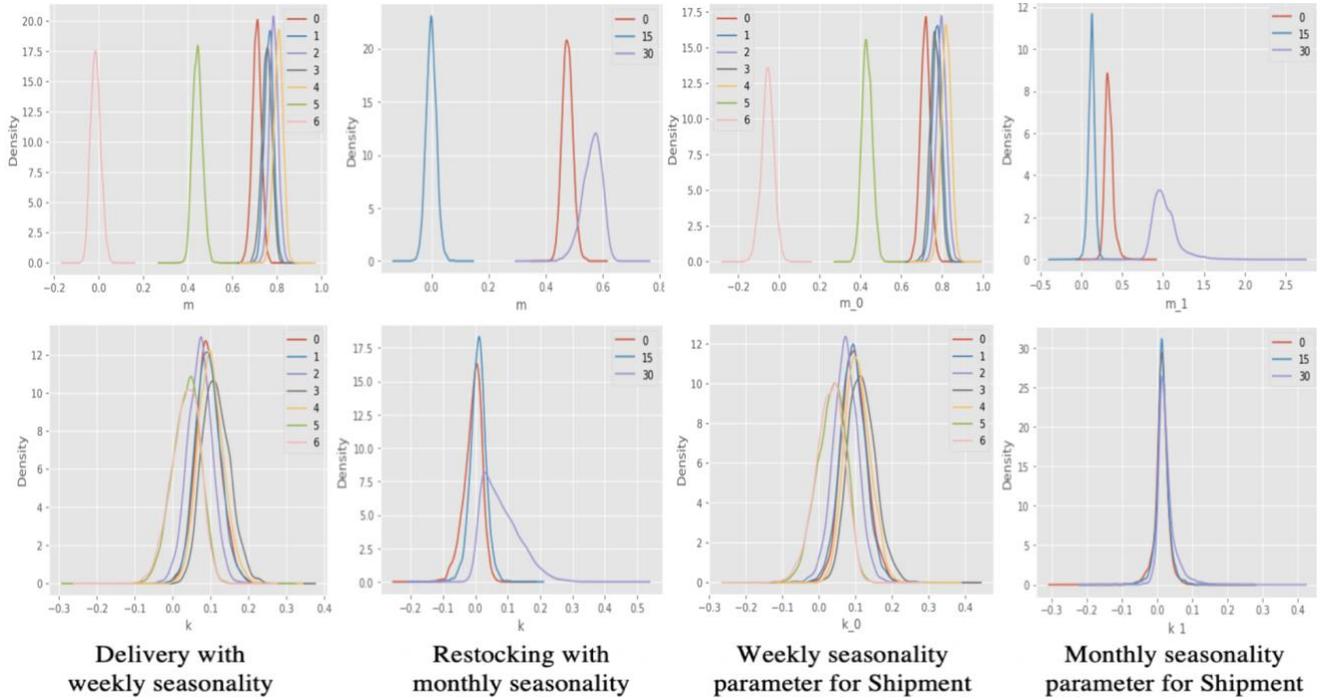

Fig. 6. Trend parameters for the delivery, restocking, and shipment datasets

5.2.2. Seasonality contribution factor

The $\theta$ parameter in equation (4c) can be used to interpret the effect of each seasonal pattern. A seasonal pattern with a higher mixture weight, theta, has a greater effect on the time series. In Fig. 7, posterior distribution of theta for weekly and monthly seasonalities are indexed with 0 and 1. For the delivery dataset where monthly seasonality is hardly observed, the weight for monthly seasonality is distributed near zero; inverse result is shown from the restocking dataset which has strong monthly seasonality. Shipment data show different pattern where both seasonality have remarkable effect as expected. The maximum a posteriori estimates of each weight are about 0.6 and 0.4, so it can be inferred that the weekly seasonality has a linear effect of about twice the monthly seasonality.

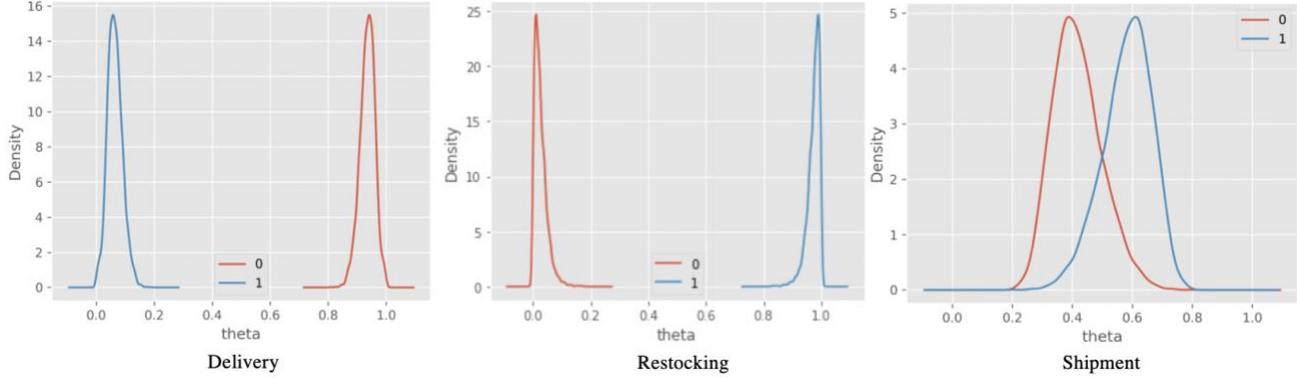

Fig. 7. Posterior distribution of seasonal factors for delivery, restocking, and shipment

## 6. Conclusions

We have proposed using MHS to develop a hierarchical model for time series data with multiple seasonal patterns. We demonstrated the applicability of the model using a real-world dataset of pallet movement from a logistic network in comparison to simpler models and models using previous methods. Through these comparisons, we confirmed that the prediction performance of our novel model in the given dataset was greatly improved. Moreover, we have shown that the fitted parameters can reasonably explain the contribution of each seasonal pattern to the time series. The enhancement of accuracy and interpretability especially in terms of seasonality would be valuable in a decision-making context. For example, based on the seasonality priority which MHS provides from theta comparison, managers could decide the optimal reorder period.

Although only the seasonal periods of week and month are introduced and modeled in this paper, other periods ranging from the minute, hour, and day to the quarter or year could be modeled using MHS. It is possible to choose different seasonal periods and to choose the number of seasonal periods to be considered compositely, meaning that the resulting model would be a true mixed hierarchical model.

We suggest three points regarding the limitation of our study. The first is trend change detection. The current model would have difficulty fitting a time series with a dynamic trend change. This could be improved by an automatic trend change detection model. This is important because time series are the combined result of trend and seasonality. An accurate trend model, as well as a clear distinction between the trend and seasonal effects, is needed for a better seasonality model. The second is limitation on application domain. Only data from logistic domain have been used to validate our model. Our proposed model focuses on time series with specific conditions: periodic scale jump and multiple seasonality. Though they are not uncommon characteristics respectively, it was not easy to find a dataset that satisfies all. We chose data from logistics because transport data had strong and multiple seasonalities. Our model improved the forecast results once it was applied to the company which provided the data; not for just the given three dataset but multiple other datasets which shows a similar pattern. However, if higher performance is observed from more diverse types of time series data, it would add to our model's advantage. Lastly, although the user interface of our model has not been discussed in this paper as it may be out of the scope, its existence would support the users who are unfamiliar with coding. In other words, by adding a user-interface, our model could be used more widely and conveniently and therefore increase its contribution in terms of an expert system.

MHS utilized the strengths of the hierarchical approach and was able to achieve both enhanced estimation accuracy and model interpretability. The new model, which addresses seasonality in a nuanced manner, could bring great benefits to business management, where seasonal patterns are salient.

**Acknowledgment**

This research was supported by the National Research Foundation of Korea (NRF) grant funded by the Korea government (MSIT: Ministry of Science and ICT) (NRF-2018R1C1B5045667).

Usher, J., & Dondio, P. (2020). BREXIT Election: Forecasting a Conservative Party Victory through the Pound using ARIMA and Facebook's Prophet. *In Proceedings of the 10th International Conference on Web Intelligence, Mining and Semantics*, 123-128.

Vehtari, A., & Lampinen, J. (2002). Bayesian model assessment and comparison using cross-validation predictive densities. *Neural computation, 14(10)*, 2439-2468.

Vehtari, A., Gelman, A., & Gabry, J. (2017). Practical Bayesian model evaluation using leave-one-out cross-validation and WAIC. *Statistics and Computing, 27(5),* 1413-1432.

Venkatesh, K., Ravi, V., Prinzie, A., & Van den Poel, D. (2014). Cash demand forecasting in ATMs by clustering and neural networks. *European Journal of Operational Research, 232(2)*, 383-392.

Weytjens, H., Lohmann, E., & Kleinsteuber, M. (2019). Cash flow prediction: MLP and LSTM compared to ARIMA and Prophet. Electronic Commerce Research, 1-21.

**Appendix A**

Complete pooling model

```
data {
   int<lower=1> T;
   vector[T] t;
   vector[T] y;
}
parameters {
   real k;
   real m;
   real<lower=0> sigma_obs;
}
transformed parameters {
   vector[T] yhat;
   yhat = k * t + m;
}
model {
   k ~ normal(0, 5);
   m ~ normal(0, 5);

   y ~ normal(yhat, sigma_obs);
```

1. Partial pooling

```
data {
   int<lower=1> T;
   int<lower=1> P;
   vector[T] t;
   vector[T] y;
   int pool[T];
}
parameters {
   real k_mu;
   real<lower=0> k_sigma;
   real m_mu;
```

```
    real<lower=0> m_sigma;
    real k[P];
    real m[P];
    real<lower=0> sigma_obs;
}
transformed parameters {
    vector[T] p_k;
    vector[T] p_m;
    vector[T] yhat;
    for (i in 1:T) {
        p_k[i] = k[pool[i]];
        p_m[i] = m[pool[i]];
    }
    yhat = p_k .* t + p_m;
}
model {
    k_mu ~ normal(0, 5);
    k_sigma ~ exponential(1);
    m_mu ~ normal(0, 5);
    m_sigma ~ exponential(1);

    k ~ normal(k_mu, k_sigma);
    m ~ normal(m_mu, m_sigma);
    sigma_obs ~ normal(0, 0.5);

    y ~ normal(yhat, sigma_obs);
}
```

2. Mixed pooling model (MHS)

```
data {
    int<lower=1> T;
    int<lower=1> D;
    int<lower=1> P;
    vector[T] t;
```

```
    vector[T] y;
    int pool[T, D];
}
parameters {
    real k_mu;
    real<lower=0> k_sigma;
    real m_mu;
    real<lower=0> m_sigma;
    real k[D, P];
    real m[D, P];
    simplex[D] theta;
    real<lower=0> sigma_obs;
}
transformed parameters {
    matrix[T, D] p_k;
    matrix[T, D] p_m;
    vector[T] yhat;
    for (i in 1:T) {
        for (d in 1:D) {
            p_k[i, d] = k[d, pool[i, d]];
            p_m[i, d] = m[d, pool[i, d]];
        }
    }
    yhat = (p_k * theta) .* t + (p_m * theta);
}
model {
    k_mu ~ normal(0, 5);
    k_sigma ~ exponential(1);
    m_mu ~ normal(0, 5);
    m_sigma ~ exponential(1);

    for (d in 1:D) {
        k[d] ~ normal(k_mu, k_sigma);
        m[d] ~ normal(m_mu, m_sigma);
    }
```

```
  sigma_obs ~ normal(0, 0.5);

  y ~ normal(yhat, sigma_obs);
}
```